\documentclass{article}
\usepackage[utf8]{inputenc}

\title{Graph based adaptive evolutionary algorithm for continuous optimization}
\author{Asmaa Ghoumari, Amir Nakib \\
Universit\'{e} Paris-Est, Laboratoire LISSI, \\ 122 Rue Paul Armangot, 94400\\ Vitry sur Seine, France\\
}

\usepackage{graphicx}

\begin{document}
\maketitle

\section{Introduction}
The greatest weakness of evolutionary algorithms, widely used today, is the premature convergence due to the loss of population diversity over generations. To overcome this problem, several algorithms have been proposed, such as the Graph-based Evolutionary Algorithm (GEA) \cite{1} which uses graphs to model the structure of the population, but also memetic or differential evolution algorithms \cite{2,3}, or diversity-based ones \cite{4,5} have been designed. These algorithms are based on multi-populations, or often rather focus on the self-tuning parameters, however, they become complex to tune because of their high number of parameters. In this paper, our approach consists of an evolutionary algorithm that allows a dynamic adaptation of the search operators based on a graph in order to limit the loss of diversity and reduce the design complexity. The algorithm uses several evolutionary operators and builds a graph that represents the possible sequences of switching between operators based on the losses or gains of diversity passing from one operator to another.

\section{Proposed Algorithm}
The objective of the proposed evolutionary algorithm is to adapt the strategies (associations of a crossing operator and a mutation operator) during the process in order to minimize the loss of diversity by choosing the best combination of operators. For that, our approach is based on a graph representing the relations between these operators. Each strategy is represented by a node, and the weights on the arcs are calculated from the measured population diversities. The strategy applied to the population will be questioned at regular intervals of a length of $\delta$ generations ($\delta$ being a parameter specific to the algorithm).

The proposed algorithm has $N = 20$ strategies at its disposal: the BLX-$\alpha$, discrete, one-point, linear, barycentric crossover operators, and the Levy, Gaussian, $DE / RAND / 1 / BIN$ and scramble mutation operators. The algorithm will challenge the strategy applied to the population regularly, every $\delta$ generations. The chosen strategy is that supposed to maximize diversity for the next $\delta$ generations. For this, we refer to the weights of the arcs of the graph, and we update them based on calculating the difference of diversities a between $\delta$ past generations, bounded between $max(0,p)$ and $min(p,0)$. If it is preserved, the weight increases by a to favour the chain of a strategy to another, otherwise the weight is decreased of a. The weight represents the probability of selection of a strategy which will be used during the strategy selection process. The latter, based on the principle of the posterior maximum, is then performed to select the operators to apply to the population for the next $\delta$ generations. Moreover, the calculation of the diversity is computed via the Euclidean distance.

\section{Results and discussions}
To evaluate the performance of the proposed algorithm, we carried out tests in dimension D = 40, and the stopping criterion was fixed to 40,000 maximum evaluations of the objective function. The other parameters being the crossover rate CR = 0.7 and the mutation rate MU = 0.3, a population size pop = 50 individuals, and $\delta= 20$ generations, N = 20 strategies. We have used 12 known optimization functions from the literature: Sphere, Schwefel 1.2, Schwefel 2.21, Griewank, Elliptic, Zakharov, Inverted cosinus mixture, Levy and Montalvo 2, Neumaier 3, Periodic, Michalewicz and Alpine. The results are calculated on 100 runs and show the best final average individual as well as the standard deviation (in parentheses) compared to the optimum in Table 1:

\begin{table}[]
\centering
\includegraphics[scale=0.70]{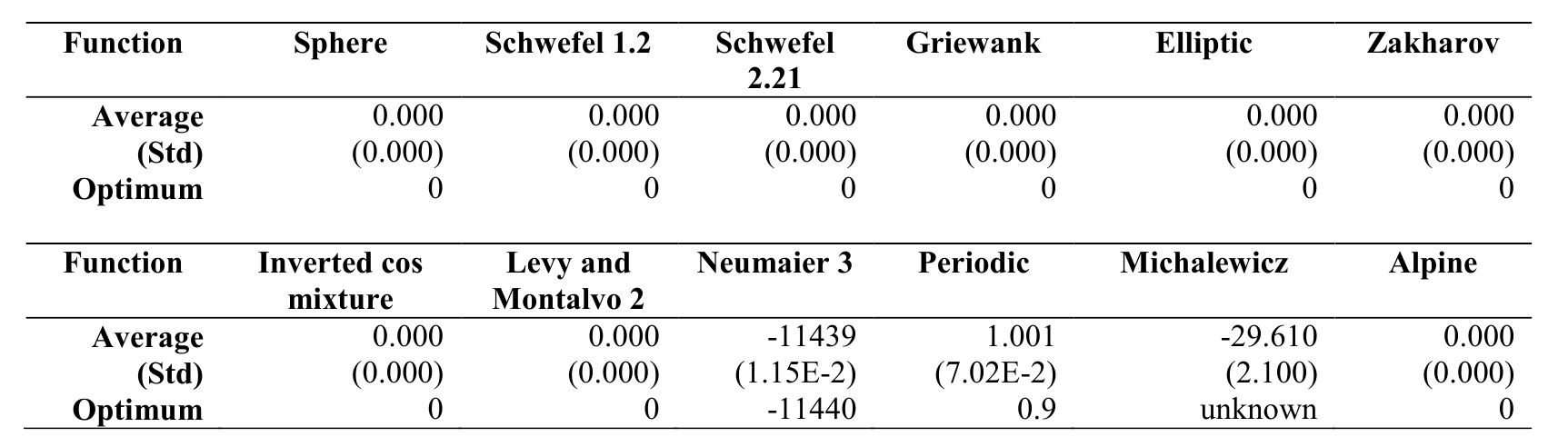}
\caption{Results in Dim=40.}
 \label{tab:my_label}
\end{table}

All test functions have their global optimum in 0, except for Periodic and Neumier 3 which are very close to the optimum. It is found that the algorithm obtains very good results for these optimization functions. However, it is difficult to reach the optimum for the Periodic and Neumier 3 functions, which are known to be particularly difficult.

Based on CEC2015, we computed the computation time of a simple evolutionary algorithm and compared it to the proposed approach in this paper. Our approach gets an average computation time more important than a classical EA, which is expected since we introduced a learning procedure to design the algorithm. However, computation time can be enhanced, easily, using parallel and distributed implementation.

Finally, the proposed approach to design an evolutionary algorithm through the use of a dynamic graph. This method allows to reduce the design effort with keeping a good optimization performance.

\begin{figure}[h!]
\centering
\includegraphics[scale=1.3]{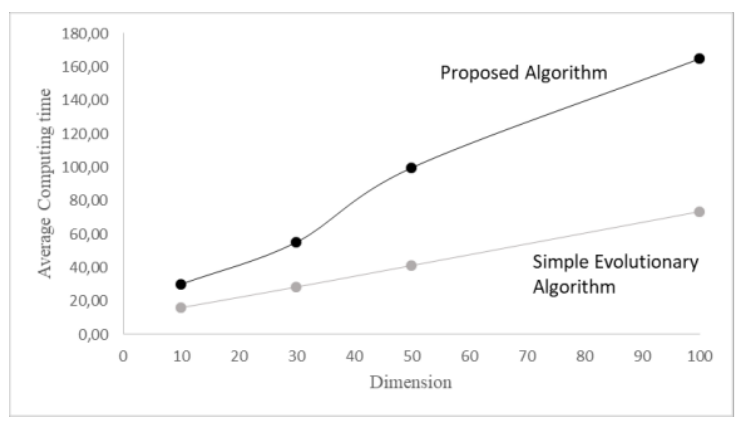}
\caption{Complexity in terms of computation time.}
\label{fig:universe}
\end{figure}

\bibliographystyle{plain}
\bibliography{main}

\end{document}